\documentclass[11pt, review,3p,sort&compress]{elsarticle}

\usepackage{amsmath,amssymb,amsfonts}
\usepackage{graphicx}
\usepackage{textcomp}
\usepackage{xcolor}

\def\BibTeX{{\rm B\kern-.05em{\sc i\kern-.025em b}\kern-.08em
		T\kern-.1667em\lower.7ex\hbox{E}\kern-.125emX}}

\usepackage[hidelinks]{hyperref}
\usepackage[utf8]{inputenc}
\usepackage{xcolor}
\usepackage{booktabs} 
\usepackage{microtype}
\usepackage{graphicx} 
\usepackage{tabularx} 
\usepackage{array}
\usepackage{paralist}
\usepackage{eurosym}
\usepackage{pbox} 
\usepackage{multirow}
\usepackage{subcaption}
\usepackage{stackengine}
\usepackage{adjustbox}
\usepackage[noend]{algpseudocode}
\usepackage{algorithm}
\usepackage{algorithmicx}
\usepackage{setspace}

\newtheorem{definition}{Definition}
\newtheorem{problem}{Problem}

\newcommand{\mypar}[1]{\smallskip\noindent\textbf{#1.}}

\newcommand{\papertitle}{Partial Order Resolution of Event Logs for Process Conformance Checking}

\journal{Decision Support Systems}

\begin{document}

\begin{frontmatter}

\title{\papertitle}

\author[mu]{Han van der Aa}
\ead{han@informatik.uni-mannheim.de}
\author[klu,hpi]{Henrik Leopold}
\ead{henrik.leopold@the-klu.org}
\author[hu]{Matthias Weidlich}
\ead{matthias.weidlich@hu-berlin.de}

\address[mu]{Data and Web Science Group, University of Mannheim Mannheim, Germany}
\address[klu]{K\"uhne Logistics University, Hamburg, Germany}
\address[hpi]{Hasso Plattner Institute, University of Potsdam, Potsdam, Germany}
\address[hu]{Department of Computer Science, Humboldt-Universit\"at zu Berlin, Berlin, Germany}

\date{Received: \today }


\begin{abstract}
		While supporting the execution of business processes, information systems 
	record event logs. Conformance checking relies on these logs to analyze 
	whether the recorded behavior of a process conforms to the behavior of a 
	normative specification. A key assumption of existing conformance checking 
	techniques, however, is that all events are associated with timestamps that allow 
	to infer a total order of events per process instance. 
	Unfortunately, this assumption is often violated in practice. Due to 
	synchronization issues, manual 
	event recordings, or data corruption, events are only partially ordered.
	In this paper, we put forward the problem of partial order resolution of 
	event logs to close this gap. It refers to the construction of a probability 
	distribution over all possible total orders of events of an instance. 
	To cope with the order uncertainty in real-world data, we present several
	estimators for this task, incorporating different notions of behavioral 
	abstraction.  Moreover, to reduce the 
	runtime of conformance checking based on partial order resolution, we 
	introduce an approximation method that comes with a bounded error in terms of 
	accuracy. Our experiments with real-world and synthetic data reveal that our 
	approach improves accuracy over the state-of-the-art  considerably. 
\end{abstract}


\end{frontmatter}

\section{Introduction}
\label{sec:introduction}

The execution of business processes is these days supported by 
information systems~\cite{DBLP:books/sp/DumasRMR18}. Whether it is the handling 
of a purchase order in 
e-commerce, the tracking of an issue in complaint management, or monitoring of 
patient pathways in healthcare, information systems track the progress of 
processes in terms of event data. An event hereby denotes the execution of a 
specific activity (e.g., checking plausibility of a
purchase order, proposing some issue resolution, creating a patient 
treatment plan) as part of a specific case (e.g., a purchase order, an issue 
ticket, a patient) at a specific point in time~\cite{DBLP:books/sp/Aalst16}. A 
collection of such 
events, 
referred to as an event log, therefore represents the 
\emph{recorded} behavior of a process.


A considerable threat to process improvement initiatives is 
\emph{non-conformance} in process execution, i.e., 
situations in which the actual behavior of a 
process deviates from the desired behavior~\cite{DBLP:books/sp/CarmonaDSW18}. Such 
differences stem from the 
fact that information systems \emph{support} process 
execution, 
 but do not 
\emph{enforce} a particular way of executing the 
process~\cite{schonenberg2008process}.
Rather, human interaction drives a process, giving people a certain 
flexibility in the execution of a particular case.
The implications of non-conformance are known to be severe. They range 
from reduced productivity~\cite{bagayogo2013impacts} to financial penalties 
imposed by authorities~\cite{lu2007compliance}. To efficiently detect cases of 
non-conformance, techniques for \textit{conformance checking}
have been introduced~\cite{DBLP:books/sp/CarmonaDSW18,caron2013comprehensive,van2011conceptual}. They strive 
for automatic detection of deviations of the recorded and desired process 
behavior, by comparing event logs with process models. 
They verify whether the causal dependencies for activity 
execution, as specified in a process model, hold true in an event log and 
provide diagnostic information on non-conformance.

However, a key assumption of state-of-the-art conformance checking techniques 
is that all events of a case are labeled with 
timestamps that allow to infer a \emph{total} 
order~\cite{adriansyah2011conformance}.
Unfortunately, this assumption is often violated. In practice, there are 
various sources affecting the quality of recorded event data, among them 
synchronization issues, manual recording of events, or unreliable data 
sensing~\cite{DBLP:conf/cidm/BoseMA13}. For instance, in healthcare 
processes, only the day of a set of treatments may be known, but not the 
specific point in time~\cite{mans2013process}. 
Hence, events are only \emph{partially} ordered, which 
renders existing conformance checking techniques inapplicable. 

In this paper, we argue that it is often possible to resolve the unknown order 
of such events. Our idea is to use information from the entire event log to 
estimate the probability of each possible total order, induced by the 
partial order of events. This way, conformance checking is grounded 
in a stochastic model, incorporating the probabilities of specific order 
resolutions. 

Our contributions and the structure
of the paper, following background on conformance checking and uncertain event 
data in the next section, are summarized as follows:

\begin{compactitem}
\item We introduce the problem of partial order resolution for event logs 
(\autoref{sec:problemstatement}). We formalize the problem and outline how it 
enables probabilistic conformance checking.
\item We present various behavioral models to address partial order resolution 
(\autoref{sec:resolution}). These models encode different levels of 
abstraction of event orders, which are then used to correlate events of 
different cases. 

\item To improve the computational efficiency of conformance checking in the presence of order uncertainty, we propose a sample-based approximation method that provides statistical guarantees on obtained conformance checking results (\autoref{sec:resultapproximation}). 

\item The accuracy and efficiency of our approach, as well as the approximation method are demonstrated through evaluation experiments 
based on  real-world and synthetic data collections (\autoref{sec:evaluation}). The conducted experiments reveal that our approach achieves considerably more accurate results than the state-of-the-art, reducing the average error by 59.0\%.

\end{compactitem}
Finally, we review our contributions in light of related
work (\autoref{sec:relatedwork}) and conclude 
(\autoref{sec:conclusion}).

\section{Background}
\label{sec:background}

\autoref{sec:conformancecheckingBG} first introduces a running example and 
illustrates the goal of conformance checking. 
\autoref{sec:uncertaineventdataBG} then discusses the impact that order uncertainty has on this task. 

\subsection{Conformance Checking}
\label{sec:conformancecheckingBG}

Conformance checking analyzes deviations between 
the recorded and the desired behavior of a process. Recorded behavior 
is given as a log of events, each event carrying at least a timestamp, 
a reference to an activity, and an identifier of the case for 
which an activity was executed. Based on the latter, a log can be 
partitioned into \emph{traces}: ordered, maximal sets of events, all related to 
the same, individual case. 

The desired behavior of a process, in 
turn, is captured by a normative 
specification, i.e., a process model. It defines causal 
dependencies for the activities of a process, thereby inducing a set of 
\emph{execution sequences}, i.e., sequences of possible activity executions that are allowed according to the process model. 
Conformance checking determines whether a recorded trace 
corresponds to an execution sequence of the process model. 
Put differently, it assesses whether a trace represents a \emph{word} of the 
\emph{language} of the process model. 

\begin{figure*}[t!]
	\centering
	\includegraphics[width=0.85\linewidth]{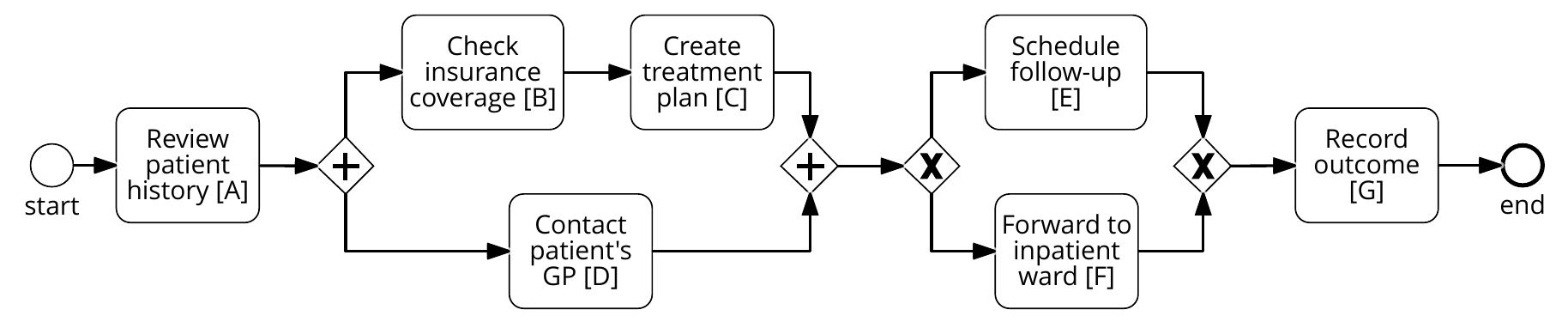}
	\vspace{-0.5em}
	\caption{Process model capturing the desired behavior for a healthcare 
		process.}
	\label{fig:exampleModel}
	\vspace{-1em}
\end{figure*}

For illustration, consider the process model depicted in 
\autoref{fig:exampleModel}, henceforth referred to as model $M$. It defines the 
desired behavior of a healthcare process, using the Business Process Model and 
Notation (BPMN). That is, in each case, a patient's history should first be 
reviewed (activity $A$), before checking their insurance coverage ($B$) and 
creating a treatment plan ($C$). In parallel, the general 
practitioner (GP) of the patient is contacted ($D$). 
Depending on the result of these 
activities, a patient either gets a follow-up appointment ($E$) or is 
forwarded to the inpatient ward ($F$), before the outcome is recorded ($G$).

Next to this process model, consider the following traces\footnote{We use 
lowercase 
letters to refer to events corresponding to a certain activity, e.g., $e$ 
describes an occurrence of activity $E$.}: $\pi_1 = \langle a, b, c, d, e, g
\rangle$, $\pi_2 = \langle a, b, c, e, d, g \rangle$, and $\pi_3 = 
\langle a, d, b, f, e, g\rangle$.
We observe that $\pi_1$ represents a proper execution sequence of the process model $M$, so we can conclude that $\pi_1$ conforms to $M$. By contrast,  $\pi_2$ and  $\pi_3$ do not. In 
$\pi_2$, activity $E$ occurs \textit{before} $D$. That is, a 
follow-up appointment was scheduled ($E$) without first contacting the patient's GP ($D$), 
even though the model explicitly specifies that these activities should occur in the reverse order. 
Trace $\pi_3$ has several different issues. First, the creation of a treatment plan~($C$) has been omitted, even though this represents a mandatory activity. 
Second, a follow-up appointment has been scheduled ($E$), even though the patient has also been forwarded to the 
inpatient ward ($F$). According to the model, these activities are defined to be mutually exclusive, therefore leading to another conformance issue.

Conformance checking techniques aim to automatically detect such deviations between recorded and desired process behavior. 
State-of-the-art techniques for this task
construct
\emph{alignments} between traces and execution sequences of a 
model to detect  
deviations~\cite{DBLP:books/sp/CarmonaDSW18,adriansyah2011conformance,munoz2014single}.
An alignment is a sequence of steps, each step comprising a pair of an event 
and an activity, or a \emph{skip} symbol $\bot$, if an event or activity is 
without counterpart. For instance, for the non-conforming trace $\pi_3$, an 
alignment with two such skip steps may be constructed with the execution 
sequence $\langle 
A,D,B,C,F,G\rangle$ of the model:
\[
\begin{array}{l c c c c c c c}
\text{Trace } \pi_3 &  a & d & b&\bot& f& e& g\\ 
\midrule
\text{Execution sequence} & A & D & B &C & F& \bot& G\\
\end{array}
\]

Assigning costs to skip steps, a cost-optimal 
alignment (not necessarily unique) is constructed for a trace in relation to 
all execution sequences of a model~\cite{adriansyah2011conformance}. 
An optimal alignment then answers not only the question whether there are 
deviations between a 
trace and the execution sequences of a model, but also enables quantification 
of non-conformance by aggregating the costs of skip steps. 
In addition, considering a log as a whole, events and 
activities that are frequently part of skip steps highlight hotspots of 
non-conformance in process execution.

\subsection{Event Data Uncertainty}
\label{sec:uncertaineventdataBG}

While various conformance 
checking techniques have been presented in recent years, they all assume and 
require event data to have been accurately recorded. 
Yet, in practice, event logs are subject to diverse quality 
issues~\cite{DBLP:conf/cidm/BoseMA13,suriadi2017event,suriadi2015event} along all the dimensions known to assess 
data quality in 
general~\cite{DBLP:books/sp/dcsa/Batini06}, e.g., accuracy, timeliness, 
precision, completeness, and reliability. 

In this work, we focus on quality issues that relate to temporal 
aspects of events, which are highly relevant in conformance checking. 
In particular, state-of-the-art conformance checking relies on 
discrete events that are assigned a precise timestamp. Hence, the commonly 
adopted notion of a trace requires all events 
of a single case to be \emph{totally} ordered by their 
timestamps~\cite{adriansyah2011conformance,munoz2014single}. 
However, this assumption is often violated in practice. Below, we illustrate 
potential reasons for this observation. While the respective phenomena may 
cause various different types of data quality issues, they particularly disturb 
the order of events as established through their timestamps.
\begin{enumerate}
	\item \textit{Lack of synchronization.} Event logs integrate data from 
	various information systems. Tracing the execution in such distributed 
	systems has to cope with unsynchronized 
	clocks~\cite{DBLP:conf/eurosys/KoskinenJ08}, making timestamps partially 
	incomparable. Also, the logical order of events 
	induced by timestamps may be inconsistent with the order of their 
	recording~\cite{mutschler2013reliable}. 

	\item \textit{Manual recording.} The execution of activities is not always 
	directly observed by information systems. Rather, people involved in process 
	execution have to record them manually. 
  Such manual recordings are subject to inaccuracies. For instance, it has been 
  observed in the healthcare domain that personnel records their work solely at 
  the end of a shift~\cite{lu2014conformance}, rendering it impossible to 
  determine a precise order of executed activities. 	
	\item \textit{Data sensing.} Event logs may be derived from sensed data as 
	recorded by real-time locating systems (RTLS). 
	Then, the construction of discrete events from raw signals is 
	inherently uncertain and grounded in probabilistic 
	inference~\cite{DBLP:conf/icde/TranSCNDS09,busany2020interval}. For instance, deriving 
	treatment events in a hospital based on RTLS positions of patients and 
	staff members does not yield fully accurate 
	traces~\cite{senderovich2016road}.  
\end{enumerate}

The above phenomena have in common that they result in imprecise event 
timestamps. In conformance checking, 
this leads to the particular problem of \emph{order uncertainty}: The 
exact order in which events of a trace have occurred is not known. 
Consider, for instance, the events shown in \autoref{tab:eventlogexample}, 
recorded for a single case of the aforementioned process. The events may have 
been captured manually, so that the timestamps only indicate the rough hour in 
which activities have been executed. Since events $b$ and $c$ carry 
the same timestamp, it is unclear whether the patient's 
insurance was checked ($B$) \textit{before} or \textit{after} creating a 
treatment plan ($C$). Since the type 
of insurance may influence the treatment plan, the model in 
\autoref{fig:exampleModel} defines an explicit execution order for both 
activities. Yet, due to order uncertainty, we cannot establish whether 
the process was indeed executed as specified.

\begin{figure}[b]
	\centering
	\hspace{2cm}
  	\begin{subfigure}{0.3\linewidth}
	\centering
	\includegraphics[width=\linewidth]{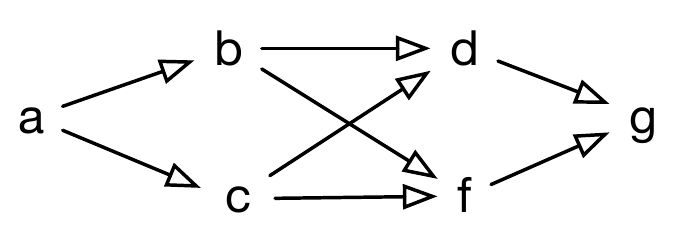}				
	\end{subfigure}%
	\hspace{1cm}
	\begin{subfigure}{0.45\linewidth}
	\phantom{.}
	 $\pi_4 = \langle a, b, c, d, f, g \rangle$ \\
	 $\pi_5 = \langle a, c, b, d, f, g \rangle$ \\
	 $\pi_6 = \langle a, b, c, f, d, g \rangle$ \\
	 $\pi_7 = \langle a, c, b, f, d, g \rangle$ \\
	\end{subfigure}
	\hfill
	\vspace{-0.5em}
	\caption{Partial order and trace resolutions resulting from the order 
	uncertainty in the case of \autoref{tab:eventlogexample}.}
	\label{fig:partialorder}
	\vspace{-1em}
\end{figure}

A result of order uncertainty, whether caused by a lack of synchronization, manual recording, or data sensing issues, is that the events of a trace are only 
\emph{partially} ordered. Such a partial order  is visualized in \autoref{fig:partialorder} for the 
events of the example case. This 
partial order induces four totally ordered sequences of events, denoted as $\pi_4$ to $\pi_7$ in the figure. 
Such a situation is highly problematic in conformance checking, because of the implied ambiguity. 
For the example in \autoref{tab:eventlogexample}, 
only one of the totally ordered event sequences, i.e., 
$\pi_4$ conforms to model $M$, whereas different kinds of deviations are detected for the 
remaining three. Hence, one cannot conclude if the case was executed as 
specified by the model at all and, if it was not, which conformance violations actually occurred. 

Without further insights into the execution of a case, two approaches 
may 
be followed to resolve order uncertainty. First, one may consider all 
induced totally ordered traces to be equally likely, so that the number of such 
traces that conform to the model provides a conformance measure. Second, 
order uncertainty may be neglected, verifying whether \emph{one} of the induced 
total orders is conforming~\cite{lu2014conformance}. As we will 
demonstrate empirically, both approaches introduce a severe bias in 
conformance checking. In this work, we therefore strive for a fine-granular 
assessment of each induced total event order. 

\begin{table}[t]
	\small
	\centering
	\caption{Events recorded for a single case.}
	\label{tab:eventlogexample}
	\begin{tabular}{c l r}
		\toprule
    \textbf{Event ID} & \multicolumn{1}{c}{\textbf{Activity}} &  
		\multicolumn{1}{c}{\textbf{Timestamp}} \\
		\midrule
		a & Review patient history [A] & 13:00 \\
		b & Check insurance coverage [B] & 14:00 \\
		c & Create treatment plan [C]& 14:00 \\
		f & Forward to inpatient ward [F] & 15:00 \\			
		d & Contact patient's GP [D]&15:00 \\
		g & Record outcome [G] & 16:00 \\
		\bottomrule
	\end{tabular}
	\vspace{-1em}
\end{table}

\section{Problem Statement}
\label{sec:problemstatement}

We first give preliminaries in terms of a formal model 
(\autoref{sec:preliminaries}), before defining the problem addressed in this 
paper, partial-order resolution (\autoref{sec:partialorderresolution}).

\subsection{Preliminaries}
\label{sec:preliminaries}

In this section, we introduce our formal model to capture normative and recorded process behavior.

\noindent \textbf{Normative process behavior.} A process model defines the execution dependencies 
between the activities of a process, establishing the normative or desired behavior. For our purposes, it is sufficient to 
abstract from specific process modeling languages (e.g., BPMN or Petri nets) and focus on the behavior 
defined by a model. 
Process models capture relations that exist among a collection of \emph{activities}. We denote the universe of such activities as $\mathcal{A}$. Then, a process model defines a set of execution 
sequences, $M\subseteq \mathcal{A}^*$, that capture sequences of activity 
executions that lead the process to its final state. 
For instance, the model in 
\autoref{fig:exampleModel} defines a total of six allowed execution sequences, including
$\langle A, B, C, D, E, G \rangle$,  $\langle A, B, C, D, E, F \rangle$, as well as variations in which activity $D$ occurs before or after $B$.

\noindent \textbf{Recorded process behavior.} The executions of activities of a process are recorded as events. If these 
activity executions happen within the context of a 
single case, the respective events are part of the same \emph{trace}. 
While most models 
define traces as sequences of events, we adopt a model that explicitly captures 
order uncertainty by allowing multiple events, even if belonging to the same 
trace, to be assigned the same 
timestamp. Hence, the events of a trace are only \emph{partially} ordered. We 
capture this by modeling traces as sequences of \emph{sets} of events, where each set contains events with identical timestamps. Captured as follows:

\begin{definition}[Traces]
\label{def:traces}
A \emph{trace} is a sequence of disjoint sets of events, $\sigma = 
\langle E_1, \dots, E_n \rangle$, with $E_\sigma = \bigcup_{1\leq i\leq n} E_i$ 
as the set of all events of $\sigma$.
\end{definition}

For a trace $\sigma = \langle E_1, \dots, E_n\rangle$, 
we refer to $E_i$, $1\leq i\leq n$, as an \emph{event set} of $\sigma$. 
This event set is \emph{uncertain}, if $|E_i| > 1$.
Accordingly, a trace that does not contain uncertain event sets is \emph{certain}, otherwise it is \emph{uncertain}. 
Intuitively, in a certain trace, a total order of events is established by the 
events' timestamps. 
For an uncertain trace, events within an uncertain event set are not ordered. 

An event log is a set of traces, capturing the events as they have been 
recorded during process execution. Moreover, for each event, we 
capture the activity for which the execution is represented by this event. The 
latter establishes a link between an event log and the activities of a process 
model. 
  
\begin{definition}[Event Log]
\label{def:event_log}
An event log is a tuple $L=(\Sigma, \lambda)$, where $\Sigma$ is a set of 
traces and $\lambda: \bigcup_{\sigma \in \Sigma} E_\sigma \rightarrow 
\mathcal{A}$ assigns activities to all events of all traces. 
\end{definition}

As a short-hand notation, we write a trace of a log not only as a 
sequence of sets of events, but also as a sequence of sets of activities. That 
is, for a trace $\langle \{e_1,e_2\}, \{e_3\} \rangle$ with $\lambda(e_1)=x$, 
$\lambda(e_2)=y$, and $\lambda(e_3)=z$, we also write $\langle \{x,y\}, \{z\} 
\rangle$.
According to this model, the case from 
\autoref{tab:eventlogexample} is captured by the trace $\sigma_1 = \langle 
\{a\},\{b,c\},\{d,f\},\{g\}\rangle$.

\subsection{The Partial Order Resolution Problem}
\label{sec:partialorderresolution}

To assess the conformance of an event log with a 
model, the order uncertainty of its traces needs to be handled. Yet, 
there may be several ways to resolve this 
uncertainty as the events of each 
uncertain event set may be ordered differently. 
We capture such different orders by means of \textit{possible resolutions} of event sets 
and, based thereon, of a trace. 

\begin{definition}[Possible Resolutions]
  \label{def:trace_resolutions}
  Given a trace $\sigma =\langle E_1, \dots, E_{n} \rangle$, we define \emph{possible resolutions} for:
  \begin{compactitem}
  \item an event set $E_{i}$, as any total order over its events, i.e.,  
  $\Phi(E_{i})=\{ 
  \langle e_1, \ldots, e_{|E_{i}|} \rangle \mid \forall\ 1\leq j,k\leq 
  |E_{i}|:  e_j~\in~E_i   \ \wedge \ 
  e_j=e_k\Rightarrow j=k\}$; 
  \item the trace $\sigma$, as any total order over events of its event sets, 
  i.e., $\Phi(\sigma)=\{ 
  \langle e^1_1, \ldots, e^{m_1}_1, \ldots , e^1_n, \ldots 
  e^{m_n}_n \rangle \mid \forall \ 1\leq i\leq n: \langle e^1_i, \ldots 
  e^{m_i}_i \rangle \in \Phi(E_{i})\}$.
  \end{compactitem}  
\end{definition}

\noindent
In the context of an event log $L=(\Sigma, \lambda)$, we lift the short-hand 
notation for traces based on the assigned activities to resolutions.
Then, for our example trace 
$\sigma_1 = \langle \{a\},\{b,c\},\{d,f,\}, \{g\}\rangle$, 
possible resolutions would be $\langle a,c,b,f,d,g \rangle$ or $\langle 
a,b,c,d,f,g \rangle$, but neither $\langle a,b,f,d,g \rangle$  (event $c$ is 
missing) nor $\langle a,c,f,b,d,g \rangle$ (events from different event sets are 
interleaved).

Although the events originally occurred in a total order, there is no way to 
recover this original order when it is obscured due to the aforementioned 
reasons for order 
uncertainty. 
However, we argue that even without identifying a single resolution, valuable 
insights on 
the conformance of a trace may be obtained. This can be achieved by assessing 
which resolutions of a trace conform and which do not conform to a process 
model. 
As illustrated above, it is crucial here to avoid basing conformance assessments purely on the 
\textit{number} of possible resolutions that conform to its associated process 
model: a single conforming resolution may be more likely to have occurred than multiple non-conforming resolutions combined.

We therefore resort to a 
probabilistic model that defines a probability distribution over a trace's 
possible resolutions. Then, conformance checking can be grounded in the 
cumulative probabilities of the resolutions that conform to a model, or show 
particular deviations, respectively. 
Following this line, a crucial problem addressed in this paper is how to assign probabilities to the possible resolutions of a partial order, which can be phrased as follows:

\begin{problem}[Partial Order Resolution]
  \label{problem}
  Let $L=(\Sigma, \lambda)$ be an event log. The \emph{partial 
  order resolution problem} is to derive, for each trace $\sigma\in \Sigma$ and 
  each possible resolution $\varphi\in 
  \Phi(\sigma)$, the probability $P(\varphi)$ of $\varphi$ representing the 
  order of event generation for the respective case. 
\end{problem}

\noindent
Based on the probabilities $P(\varphi)$ of each resolution 
$\varphi\in \Phi(\sigma)$, 
the probabilistic conformance of a trace $\sigma$ with respect to a model $M$ can be assessed. 
Let $\mathit{conf}(\varphi, M)$ be a function that quantifies the conformance 
of a resolution to model $M$. Exemplary functions are a binary function 
$\mathit{conf}_{\mathit{bin}} : \varphi \times M \rightarrow \{0, 1\}$, 
indicating a conforming resolution with 1 and a non-conforming with 0, or a 
function based on trace fitness~\cite{DBLP:books/sp/CarmonaDSW18}, yielding a range 
from 0 to 1, i.e., $\mathit{conf}_{\mathit{fit}} : \varphi \times M \rightarrow 
[0, 1]$. Based on such a function, the weighted conformance of a trace is 
denoted as follows: 

\begin{equation}
\label{eq:probconf}
\mathit{P_{\mathit{conf}}}(\sigma, M) = \sum_{\varphi \in \Phi(\sigma)} P(\varphi) \times \mathit{conf}(\varphi, M)
\end{equation}

\noindent Similarly, more fine-granular feedback based on non-conformance may be given by analyzing alignments obtained per possible resolution of a trace. 
For instance, the probabilities assigned to resolutions can be incorporated as 
weighting factors in the aggregation of the non-conformance measured per 
possible resolution. This manifests itself in the form of the accumulative 
probabilities associated with the skip steps ($\bot$) in an alignment,  as 
shown in \autoref{sec:conformancecheckingBG}. In this way, probabilistic 
conformance checking can be used to identify hotspots of non-conformance in 
processes.

\section{Partial Order Resolution}
\label{sec:resolution}

To estimate the probability of a partial order resolution, 
we follow the idea that the context of a trace provided by the event log is 
beneficial. A business process is structured through the causal dependencies 
for the execution of activities. These dependencies are 
manifested in the traces in terms of behavioral regularities. 
Assuming that order uncertainty occurs independently of the execution of the 
process, behavioral regularities among the traces may be exploited for 
partial order resolution. The probability of a specific resolution for a given 
trace may be assessed based on order information derived from similar traces 
contained in the event log. Intuitively, if one possible resolution denotes an 
order of activity executions that is frequently observed for other 
 traces, this resolution is expected to be more likely than another 
resolution that denotes a rare execution sequence. 
It is important to note that model characteristics cannot be leveraged to determine
the likelihood of a resolution since this would introduce a bias towards conforming resolutions. 

Using traces for partial order resolution requires a careful selection of the abstraction 
level based on which traces are compared. In practice, event logs contain 
traces encoding a large number of different sequences of activity 
executions. Reasons for that are concurrent execution of activities, which 
leads to an exponential blow-up of the number of execution sequences, as well 
as the presence of noise, such as incorrectly recorded events. 
For a possible resolution of a trace, it may 
therefore be impossible to observe the exact same sequence of activity 
executions in another trace, unaffected by order uncertainty. 
We cope with this issue by defining behavioral models that realize different 
levels of abstraction in the comparison of traces. We propose 
(i) the trace equivalence model, (ii) the N-gram model, and (iii) the weak 
order model, see \autoref{tab:probabilisticmodels}. 
The models differ in the notion of 
behavioral regularity that is used for the partial order resolution.

\begin{table}[t]
	\centering
	\small
	\caption{Proposed behavioral models.}
	\label{tab:probabilisticmodels}
	\begin{tabular}{l@{\hspace{1em}}c@{\hspace{1em}}l}
  \toprule
	\textbf{Model} & \textbf{Illustration} & 
	\multicolumn{1}{c}{\textbf{Basis}} \\
  \midrule
& \multirow{3}{*}{\includegraphics[width=0.14\linewidth, trim={2mm 0mm 2mm 
2mm},clip]{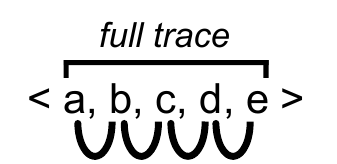}} 
&  \\
Trace equivalence  & & Equal, certain traces\\
&  \\
	
		\midrule
		& \multirow{3}{*}{\includegraphics[width=0.14\linewidth,trim={2mm 0mm 2mm 
        2mm},clip]{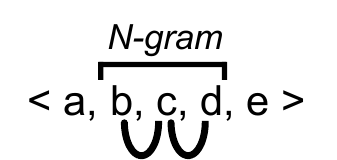}} &  \\
		N-gram &  & Equal sub-sequences of length $N$	 \\
&  \\
				\midrule		
		& \multirow{2}{*}{\includegraphics[width=0.14\linewidth,trim={2mm 0mm 2mm 
        2mm},clip]{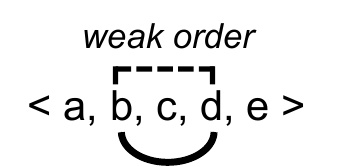}} &  \\
		Weak order &  & Indirectly follows relation of events \\
&  \\
  \bottomrule
 \end{tabular}
 \vspace{-1em}
\end{table}

\subsection{Trace Equivalence Model}
\label{sec:traceequivalence}

This model estimates the probability of a resolution by exploring how often the 
respective sequence of activity executions is observed in the event log,
in traces without order uncertainty.
To this end, we first clarify that two resolutions shall be considered 
to be equivalent, if they represent the same sequences of 
executed activities.

Let $L=(\Sigma, \lambda)$ be an event log and $\sigma,\sigma'\in \Sigma$ two 
traces of the same length, i.e., $|E_{\sigma}|=|E_{\sigma'}|$. Let $\varphi=\langle 
e_1,\ldots, e_n\rangle \in \Phi(\sigma)$ and 
$\varphi'=\langle e'_1,\ldots, e'_n\rangle \in 
\Phi(\sigma')$ two of their resolutions. Then, the resolutions are 
equivalent, denoted by $\varphi\equiv\varphi'$, if and only if $\lambda(e_i) = 
\lambda(e'_i)$ for $1\leq i\leq n$. 

We define $\Sigma_{certain} = \{ \sigma \in \Sigma \mid |\Phi(\sigma)| = 1 \}$ 
as the set of all certain traces, i.e., all traces that do not have uncertainty and, thus, only a single resolution. 
Then, we quantify the 
probability associated with a resolution $\varphi\in \Phi(\sigma)$ of a trace 
$\sigma$ as the fraction of certain traces for which the resolutions are 
equivalent:
\begin{equation}
\label{eq:fulltraceequivalence}
P_{trace}(\varphi) = \frac{ \left|\left\{ \sigma \in \Sigma_{certain} \mid 
\exists \ 
\varphi'\in \Phi(\sigma): 
\varphi' \equiv\varphi \right\}\right|}{|\Sigma_{certain}|}
\end{equation}

The above model enables a direct assessment of the 
probability of a resolution. Yet, it may have limited applicability: (i) 
It only considers certain traces, and there may only be a small number of those 
in a log; (ii) none of the certain traces may show an equivalent 
resolution, as it requires the \emph{entire} sequence of activity executions to 
be the same.

\subsection{N-Gram Model}
\label{sec:ngram}

As a second approach, 
we introduce a behavioral model based on \textit{N-gram approximation}. 
It takes up the idea of sequence approximations as they are employed in a 
broad range of applications, such as 
prediction~\cite{chen2012playlist} and speech 
recognition~\cite{mohri2002weighted}. Specifically, N-gram approximation 
enables us to define a more abstract notion of behavioral 
regularities that determines which traces shall be considered when computing 
the probability of a resolution. 

Given a resolution of a trace, 
this model first estimates the probability of 
the individual events of the resolution occurring at their specific position. 
Here, up to $N-1$ events preceding the respective event are 
considered and their probability of being followed by the event in question is 
determined. The latter is based on all  traces of the log that comprise 
sub-sequences of the same activity executions without order uncertainty. 
For instance, for $N=4$ and a resolution $\varphi_1 = \langle 
a,b,c,d,f,g \rangle$, we determine the likelihood that event $f$ occurs at the fifth position by exploring the likelihood that a sequence $\langle b, c, d 
\rangle$ is followed by  $f$. This estimation is based on all 
traces of the log that comprise events of the sequence $\langle b, c, d, f \rangle$ without order uncertainty.

To formalize this idea, we first define a predicate \emph{certain}. 
Given a log $L=(\Sigma, \lambda)$, this predicates holds for a sequence of 
activities $\langle a_1, \ldots, a_{m} 
\rangle$, $a_i\in \mathcal{A}$ for $1\leq i\leq m$ and a 
trace $\sigma = \langle E_1, \ldots, E_n\rangle \in \Sigma$, $m\leq n$, if 
$\sigma$ 
contains events for the respective activity executions without order 
uncertainty:

\vspace{-1.6em}
\begin{equation}
\begin{split}
certain(\langle a_1, \ldots, a_{m}\rangle, \sigma = \langle E_1, \ldots, 
E_n\rangle) \Leftrightarrow & \\
\exists \ i\in \{0,\ldots,n-m\},  \forall \ j\in \{1,\ldots,m\}:
E_{i+j}=\{e_{i+j}&\} \wedge \lambda(e_{i+j})=a_j.\\
\end{split}
\end{equation}
\vspace{-0em}

Using this predicate, we define the probability of events related to activity 
$a$ to follow events denoting the execution of some activities $\langle a_1, 
\dots, a_m \rangle$. This 
definition is based on the number of times the two respective sequences, with 
and without $a$, are observed in the traces of the event log:

\begin{equation}
\label{eq:ngramapproximation1}
P(a \mid \langle a_1, \dots, a_m \rangle) = \frac{\left|\left\{\sigma \in 
\Sigma\mid 
certain(\langle a_1, \dots, a_m, a \rangle, \sigma)  \right\}\right|}{
\left|\left\{\sigma \in 
\Sigma\mid 
certain(\langle a_1, \dots, a_m\rangle, \sigma)  \right\}\right|}.
\end{equation}

For illustration, we return to the example of estimating the probability 
of events related to $f$ to be preceded by those 
representing the execution of activities $\langle b, c, d \rangle$. That is, we 
divide the number of occurrences of $\langle b, c, d, f \rangle$ by the number 
of occurrences of $\langle b, c, d \rangle$, while considering only traces that 
do not show order uncertainty for the respective events.
Based thereon, the probability of a resolution is derived by 
aggregating the N-gram-based probabilities of all its events: 

\vspace{-1.6em}
\begin{equation}
\label{eq:ngramapproximation2}
P_{N\text{-}gram}(\varphi=\langle e_1, \dots e_n \rangle) =  
\prod\limits_{k=2}^n P(\lambda(e_k) \mid \langle \lambda(e_{\max(1,k-N+1)}), 
\dots, 
\lambda(e_{k-1}) \rangle) 
\end{equation}

The above approach may be adapted to explicitly consider the first events of 
traces in the assessment. Technically, an artificial event is added to the 
beginning of all traces, so that it will be part of the respective N-gram 
definitions. For instance, the example trace $\sigma_1 = \langle 
\{a\},\{b,c\},\{d,f,\}, \{g\}\rangle$ would be changed to 
$\sigma'_1 = \langle\{\circ\}, \{a\},\{b,c\},\{d,f,\}, \{g\}\rangle$ with 
$\circ$ denoting the start of the trace. Then, the estimation of the 
probability for the resolution $\varphi'_1 = \langle \circ,a,b,c,d,f,g \rangle$ 
would, using $N=4$, be based on an assessment of the probability that $c$ is 
preceded by $\langle \circ, a, b\rangle$. This explicitly considers solely 
traces that start with events that denote executions of $a$ and $b$. 

Compared to the trace equivalence model, the N-gram model is more abstract. 
This makes the model more generally applicable, as it requires only the 
presence of traces that show equivalent sub-sequences of activity executions without order uncertainty, instead of requiring fully equivalent traces without order uncertainty.  
This is particularly useful to identify local dependencies that are independent of other choices in a process. For example, looking back at the running example of \autoref{fig:exampleModel}, a 2-gram model would  clearly be able to learn the ordering dependency between checking a patient's insurance coverage (activity $B$) and establishing a treatment plan ($C$). However, this probability would not be dependent on whether activity $C$ is eventually followed by scheduling a follow-up ($E$) or by forwarding a patient ($F$). 
As such, the N-gram model would be able to identify this order requirement while imposing fewer requirements on the available data than the trace equivalence model.
The parameter $N$, furthermore, provides further flexibility. 
Higher values of $N$ lead to longer sub-sequences being considered, which 
induces a stricter notion of behavioral regularities to be exploited in 
partial order resolution. Lower values, in turn, decrease this strictness, 
thereby increasing the amount of evidence on which the resolution is based.

Yet, the N-gram model assumes that behavioral regularities materialize in the 
form of consecutive activity executions. Even if 
$N=2$, only events that (certainly) follow upon each other directly in a trace 
are considered in the assessment.

\subsection{Weak Order Model}
\label{sec:weakorder}

To obtain an even more abstract model, we drop 
the assumption that behavioral regularities relate solely to consecutive 
executions of activities. Rather, indirect order dependencies, referred to as a 
\emph{weak 
order}, among the activity executions, and thus events, are exploited. 

To illustrate this idea, consider two resolutions,  
$\varphi_1 = \langle a,b,c,d,f,g \rangle$ and $\varphi_2= \langle a,c,b,d,f,g 
\rangle$, of our example trace~$\sigma_1$. To estimate their 
probabilities, under a weak order model, 
we determine the fraction of traces in which an event related to activity 
$b$ occurs at some point before 
$c$, or vice versa, to obtain evidence about the most likely order. 
Assume that the event log also contains an uncertain trace $\sigma_2=\langle 
\{a,b\},\{d\},\{c\}, \{f,g\}\rangle$. In this trace, 
$b$ and $c$ are not part of consecutive event sets, and $b$ is even part of an 
uncertain event set. Still, this trace provides evidence that activity $b$ is 
executed before $c$, which supports resolution $\varphi_1$ of trace 
$\sigma_1$. Thereby, the weak order model would appropriately gain evidence that 
a patient's insurance coverage  should be checked ($b$) before establishing a treatment plan ($c$). 
Similarly, this model enables us to incorporate information from consecutive 
uncertain event sets, as in $\sigma_1=\langle 
\{a\},\{b,c\},\{d,f\}, \{g\}\rangle$. Due to order uncertainty, it is unclear 
whether events $c$ and $f$ directly followed each other. However, $c$ definitely occurred 
earlier than $f$, i.e., a treatment plan was certainly created ($c$) before forwarding the patient to an inpatient ward ($f$).
 Such information would not be taken into account by the 
N-gram model.

Formally, let $L=(\Sigma, \lambda)$ be an event log and $a,a'\in \mathcal{A}$ 
two activities. We define a predicate \emph{order} to capture whether a trace 
comprises events representing the 
executions of these activities in weak order:

\vspace{-2.6em}
\begin{equation}
order(a, a', \sigma = \langle E_1, \ldots, 
E_n\rangle) \Leftrightarrow 
\exists \ i,j\in \{1,\ldots, n\}, i<j: e_i \in E_i \wedge e_j \in E_j \wedge 
\lambda(e_i)=a \wedge \lambda(e_j)=a'.\\
\end{equation}

This predicate enables us to estimate the probability of having events 
related to specific activity executions in weak order. More specifically, we 
determine the ratio of traces that contain the respective events:

\vspace{-0.6em}
\begin{equation}
P(a, a') = \frac{|\{ \sigma\in \Sigma \mid order(a,a',\sigma)\}|}{|\{\sigma\in 
\Sigma  \mid \exists \ e,e' \in E_\sigma: \lambda(e)=a \wedge 
\lambda(e')=a'\}|}.
\end{equation}  

Based thereon, the probability of a resolution is defined by aggregating the 
probabilities of all pairs of events to occur in the particular order:

\begin{equation}
P_{WO}(\varphi=\langle e_1, \dots e_n \rangle) =  
\prod\limits_{\substack{1\leq i< n\\i<j\leq n}} P(\lambda(e_i), 
\lambda(e_j)).
\end{equation}

Compared to the other two models, the weak order model employs the most 
abstract notion of behavioral regularity when resolving partial orders. 
Consequently, the computation of the probability of a particular resolution can 
exploit information from many traces of an event log.

\section{Result Approximation}
\label{sec:resultapproximation}
A key issue hindering the applicability of conformance checking 
techniques in industry is their computational 
complexity, since state-of-the-art algorithms suffer from an exponential 
runtime complexity in the size of the process model and the length of the 
trace~\cite{DBLP:books/sp/CarmonaDSW18}. In the 
context of this paper, this is particularly problematic: Order uncertainty 
exponentially increases the number of conformance 
checks that are required per trace. To increase the applicability of 
conformance checking under order uncertainty, this section therefore proposes 
an approximation method incorporating statistical guarantees. 

This section discusses the calculation of expected conformance values (\autoref{sec:expectedconformance}) and associated confidence intervals (\autoref{sec:confidenceintervals}), before describing approximation method itself (\autoref{sec:computationmethod}).

\subsection{Expected Conformance}
\label{sec:expectedconformance}
To reduce the computational complexity of the conformance checking task, we 
propose an approximation method that provides statistical guarantees about the 
conformance results $P_{\mathit{conf}}(\sigma, M)$ obtained for a trace 
$\sigma$. The method provides a confidence interval for the conformance 
results, which is computed  based on conformance checks performed for a sample 
of its  possible resolutions $\Phi(\sigma)$. We use $\bar{\Phi} \subseteq 
\Phi(\sigma)$ to denote this sample and $\bar{p} = \sum_{\varphi \in 
\bar{\Phi}} P(\varphi)$ to denote the cumulative probability of the resolutions 
in the sample. Then, we define the expected conformance for a trace $\sigma$ to 
a process model $M$ as follows:

\begin{equation}
	\label{eq:estimatedconformance}
	E(P_{\mathit{conf}}(\sigma, M)) =  \sum_{\varphi \in \bar{\Phi}} P(\varphi) 
	\times \mathit{\mathit{conf}}(\varphi, M) + (1 - \bar{p}) \times 
	\mu_{\mathit{conf}}
\end{equation}

\noindent \autoref{eq:estimatedconformance} consists of two components: (i) the 
known, weighted conformance of the sampled resolutions, i.e., $\sum_{\varphi \in 
\bar{\Phi}} P(\varphi) \times \mathit{conf}(\varphi, M)$, and (ii) an estimated 
part,
$(1 - \bar{p}) \times \mu_{conf}$. This estimated part receives a weight  of 
$1- \bar{p}$, i.e., the cumulative probability of the traces not included in 
the sample. The estimate itself, i.e., $\mu_{\mathit{conf}}$, reflects the 
expected conformance of a previously unseen resolution. This 
value is obtained by fitting a statistical 
distribution over the conformance values obtained for the sample $\bar{\Phi}$. 
Consider the conformance functions discussed in 
\autoref{sec:partialorderresolution}: For a binary function  
$\mathit{conf}_{\mathit{bin}}$ with range $\{0, 1\}$, the 
results of a sample represent a \textit{Binomial distribution}. For a more 
fine-granular function $\mathit{conf}_{\mathit{fit}}$ based on 
trace fitness with range $[0, 1]$, the resulting distribution can be 
characterized using a \textit{normal distribution} for a sufficiently large 
sample (e.g., size over 20) following the central limit 
theorem~\cite{rosenblatt1956central}.

For illustration, consider a sample $\bar{\Phi}$ that contains 30 
resolutions, 21 conforming and 9 non-conforming. Then, the estimated, binary 
conformance of unseen resolutions is given as $\mu_{\mathit{conf}} = 0.70$. 
With a cumulative probability of $\bar{p} = 0.80$, the estimated component of 
\autoref{eq:estimatedconformance} equals $(1 -0.80) \times0.70 = 0.14$. Note 
that determining the expected conformance $\mu_{\mathit{conf}}$ is independent 
of the probabilities assigned by a behavioral model. Therefore, due to 
differences among the probabilities associated with the resolutions in 
$\bar{\Phi}$, it does not necessarily hold that  $E(P_{\mathit{conf}}(\sigma, 
M)) = \mu_{\mathit{conf}}$.  For instance, for the given example, we may have $ 
\sum_{\varphi \in \bar{\Phi}} P(\varphi) \times \mathit{\mathit{conf}}(\varphi, 
M) = 0.60$, yielding an estimated overall conformance of $0.60 + 0.14 = 0.74$, which is considerably higher than $\mu_{\mathit{conf}}$.  

\subsection{Confidence Intervals}
\label{sec:confidenceintervals}

Based on statistical distributions established for expected conformance 
values, we further derive statistical bounds in the form of a confidence 
interval for the estimation of $P_{\mathit{conf}}(\sigma, M)$. Recognizing that 
one part of the conformance checking results is known, whereas the other 
requires estimation, we define the confidence interval as 
follows:

\begin{equation}
\label{eq:conformancebound}
CI_{\alpha} = E(P_{\mathit{conf}}(\sigma, M))  \pm (1 - \bar{p}) \times 
m_{\alpha}
\end{equation}

\noindent \autoref{eq:conformancebound} consists of two components: (i) the expected 
conformance $E(P_{\mathit{conf}}(\sigma, M))$, and (ii) a margin, $\pm (1 - 
\bar{p}) \times m_{\alpha}$, which determines the size of the interval. Here, 
$m_{\alpha}$ denotes the \textit{margin of error} of the distribution for a 
significance level $\alpha$. The margin of error  
for a Binomial distribution obtained over a set of binary conformance 
assessments, $m_\alpha$ is computed using, e.g., the \textit{Wilson score 
interval}~\cite{agresti1998approximate}. For a normal distribution, the margin 
of error is based on the standard error~\cite{lohr2019sampling}.  

It is important to note that the width of a confidence interval established 
using \autoref{eq:conformancebound} decreases if the cumulative probability of 
the sample ($\bar{p}$) is higher. This property naturally follows, because the 
margin of error $m_{\alpha}$ is only applicable to the estimated component of a 
conformance assessment, which has the weight $1 - \bar{p}$. 
We utilize this 
property in the method described next.

\subsection{Computation Method}
\label{sec:computationmethod}

Our proposed method for efficient conformance approximation is presented in 
\autoref{alg:conformanceapproximation}. The approximation is an iterative 
procedure that incorporates the conformance of newly sampled resolutions until 
a sufficiently accurate conformance value is established. 

\begin{center}
	\begin{minipage}{.8\linewidth}
		\begin{algorithm}[H]
			\caption{Statistical conformance approximation method}
			\label{alg:conformanceapproximation}
			\small 
			\begin{algorithmic}[1]
				\State{\textbf{input} $\sigma$, a trace; $M$, a process model; $B$, a 		behavioral model; $\mathit{conf}$, a conformance } 
				\State{\phantom{\textbf{input}} 
					function; $\alpha$, a significance level; $\delta$, a desired accuracy threshold.}
				\State{$\bar{\Phi} \leftarrow \emptyset$} \Comment{The set of sampled 
					resolutions}
				\State{$R \leftarrow [\, ]$} \Comment{Bag of conformance results}
				\State{$\bar{p} \leftarrow 0$} \Comment{Accumulated probability of sampled 
					resolutions}
				\Repeat
				\State{$  \varphi  \leftarrow \mathit{select}( \Phi(\sigma) \setminus 
					\bar{\Phi}, B) $} \Comment{Sample a new resolution}  \label{line:select}
				\State{$\bar{\Phi} \leftarrow \bar{\Phi} \cup \{\varphi\} $} \Comment{Add 
					resolution to sample}
				\State{$R\leftarrow R \uplus [\mathit{conf}(\varphi, M)]$} \Comment{Compute 
					and add 
					conf. result} \label{line:confcheck}
				\State{$\mathcal{D} \leftarrow \mathit{fit}(R)$} \Comment{Fit  
					distribution on the results} \label{line:fit}
				\State{$E \leftarrow \mathit{estimateResult}(R, \mathcal{D})$} \Comment{Use 
					\autoref{eq:estimatedconformance}} \label{line:resultestimate}
				\State{$m_{\alpha} \leftarrow \mathit{computeMargin}(\mathcal{D, \alpha})$} 
				\Comment{Margin of error} \label{line:margin}
				\State{$\bar{p} \leftarrow \bar{p} + P_B(\phi)$} 
				\Comment{Increase accumulated probability} \label{line:increment_p}
				
				\Until{$(1 - \bar{p}) \times m_{\alpha} / E \leq \delta$ } \Comment{Check 
					threshold} \label{line:check}
				\State{\textbf{return} $E \pm m_{\alpha}$} \Comment{Return confidence 
					interval}
			\end{algorithmic}
		\end{algorithm}
	\end{minipage}
\end{center}

\textbf{Loop.}
Each iteration starts by selecting a resolution $\phi$, which has a maximal 
likelihood from those in $\Phi(\sigma) \setminus \bar{\Phi}$ 
(line~\ref{line:select}). This selection is guided by one of the behavioral 
models introduced in \autoref{sec:resolution}, as configured by the input 
parameter $B$. 
Next, the algorithm computes the conformance result for $\phi$ and adds it to 
$R$, the bag of conformance results (line~\ref{line:confcheck}). Then, 
statistical distribution is fit to the results 
sample (line~\ref{line:fit}). Based on this distribution, the estimated result 
is computed using \autoref{eq:estimatedconformance} 
(line~\ref{line:resultestimate}), before the margin of error is determined 
(line~\ref{line:margin}) and the accumulated probability of all sampled 
resolutions is updated (line~\ref{line:increment_p}).

\textbf{Stop condition.} The iterative procedure is repeated until the margin 
of error leads to results that are below a user-specified precision threshold 
$\delta$. In particular, the method samples resolutions 
until the ratio of the margin of error $m_{\alpha}$ and the estimated 
conformance $E$, weighted by the complement of the accumulated probability of 
sampled resolutions, is below $\delta$ (line~\ref{line:check}). 
The stop condition is based on the ratio, rather than the absolute margin of error, given that, for instance, a margin of $0.05$ has a considerably greater impact when $E = 0.2$ than compared to $E = 0.8$. 

By employing this approximation method, we obtain results that satisfy a 
desired significance value $\alpha$ and precision level $\delta$. This means 
that, when possible, the method requires only a relatively low number of 
resolutions, whereas for traces with a higher variability among its 
resolutions, the method will use a greater sample to ensure result accuracy.

\section{Evaluation}
\label{sec:evaluation}
This section describes evaluation experiments in which we assess the accuracy of the proposed behavioral models for conformance checking. 
We achieve this by comparing the conformance checking results obtained for 
uncertain traces to the conformance results that would have been obtained 
without order uncertainty. These results are also compared against two 
baselines.
Both the datasets and the implementation used to conduct these experiments are publicly available.\footnote{\url{https://github.com/hanvanderaa/uncertainconfchecking}}

\subsection{Data}
\label{sec:evaluation:data}

We conducted our evaluation based on both real-world and synthetic data collections. 

\textbf{Real-world collection.} We used three, publicly available, real-world 
events logs, detailed in \autoref{tab:realworldcollectionCharacteristics}. As 
shown in the table, the three logs differ considerably, for instance in terms 
of their size (13,087 for BPI-12 to 150,370 traces for the traffic fines log) 
and trace length (averages from 3.7 to 20.0 events per trace). 
These logs also demonstrate the prevalence of coarse-grained timestamps in 
real-world settings, since all logs contain a considerable amount of uncertain 
traces, ranging up to 93.4\% of the traces for the BPI-14 log. 
To obtain a process model that serves as a basis for conformance checking, we use the inductive miner~\cite{leemans2013discovering}, a state-of-the-art process discovery 
technique, with the default 
parameter setting (i.e., a noise filtering threshold of 80\%).

\begin{table}[!ht]
\centering
\small
\setlength{\tabcolsep}{0.7em}
\caption{Characteristics of the real-world collection.}
\label{tab:realworldcollectionCharacteristics}
\begin{tabular}{llll}
	\toprule
	\textbf{Characteristic } & \textbf{BPI-12~\cite{bpi2012}} & 
	\textbf{BPI-14~\cite{bpi2014}} &\textbf{Traffic fines~\cite{trafficfines}}  \\
	\midrule 
	Places & 32 & 27 & 23 \\
	Transitions & 45 & 29 & 26 \\
	Traces & 13,087 & 41,353 & 150,370 \\
	Variants & 4,366 & 31,725 & 231\\
	Trace length (avg.) & 20.0 & 7.3 & 3.7 \\
	Uncertain traces & 5,006 (38.3\%) & 38,649 (93.4\%) & 9,166 (6.1\%) \\
	Events & 262,200 & 369,485 & 561,470\\
	Events in uncertain event sets & 25,369 (9.7\%) & 224,515 (60.7\%) & 21,308 (3.8\%) \\
	Resolutions\hphantom{..}(avg.) & 21.8 & 91.7 & 3.2 \\
	Resolutions (max.) & 3,072 & 4,608 & 12 \\
	\bottomrule		
\end{tabular}
\end{table}

\textbf{Synthetic collection.} We have generated a collection of 500 synthetic models with varying characteristics, including aspects such as loops, arbitrary skips, and non-free choice constructs.
By using synthetic logs, we are able to assess the impact of factors such as the degree of non-conformance and order uncertainty on the conformance checking accuracy of our approach.
We employed the state-of-the-art process model generation technique 
from~\cite{jouck2018generating} due to its ability to also generate 
non-structured models, e.g., models that include non-local decisions. 
To generate models, we employed 
the default parameters set by the technique's developers.

For each of these models, we used 
a stochastic simulation plug-in of the ProM~6 
framework~\cite{rogge2013prediction} to generate an event log consisting of 
1000 traces, using the default plug-in settings (i.e., uniform probabilities for all choices and exponential inter-arrival and execution times). The main characteristics of 
these models and their logs are detailed in 
\autoref{tab:syntheticCollection}. 

To introduce non-conformance, we insert noise using the same simulation plug-in, which randomly inserts, swaps, and removes events in a specified fraction of the traces. For each model, we created logs with four different noise levels, by inserting noise into 25\%, 50\%, 75\%, and 100\% of the traces. This results in four sets of logs with, on average 353.8, 564.0, 774.2 and 998.4 non-conforming traces, respectively. 
We introduced \textit{ordering uncertainty} by abstracting all timestamps to minutes, i.e., by omitting all information on seconds and milliseconds. As a result, about 50.6\% of the traces in the event logs have some order uncertainty in the form of at least two unordered events. The uncertain traces 
have an average of 4.0 possible resolutions per  trace, up to a maximum of 16,384.

\begin{table}[!t]
	\caption{Characteristics of the synthetic collection.}
	\label{tab:syntheticCollection}
	\centering
	\small
	\begin{tabular}{lrrc@{\hspace{3em}} lrr}
		\cline{1-3} \cline{5-7}
		\noalign{\smallskip}
		\multicolumn{3}{c}{\textbf{Per process model}} && \multicolumn{3}{c}{\textbf{Per event log}} \\
		\textbf{Node Type} &  \textbf{Avg.} & \textbf{Max.} && \textbf{Characteristic} &  \textbf{Avg.} & \textbf{Max.} \\
		\noalign{\smallskip}
				\cline{1-3} \cline{5-7}
				\noalign{\smallskip}
		Places &   19.1 & 77  &	&Traces & 1,000.0 & 1,000 \\
		Transitions &19.2 & 84 & & 			 Trace length & 6.4 & 122 		\\
		
		And-splits& 4.8 & 54 && Uncertain traces& 506.4 & 873	\\	
		Xor-splits& 3.4 & 28 &&	Events in uncertain event sets & 20.1\%  & 68.7\%\\
		Silent steps & 6.9 & 64 &&  Resolutions & 4.0 & 16,384 \\
		\noalign{\smallskip}
		\cline{1-3} \cline{5-7}
	\end{tabular}
\end{table}

The model-log pairs included in the real-world and synthetic collections differ considerably in terms of model complexity, trace length, degree of non-conformance, and degree of uncertainty. As a result, these collections enable us to assess the impact of such key factors on the conformance checking accuracy and efficiency of our proposed behavioral models and approximation method. In this way, we make sure the experimental results have a sufficiently high level of external validity.

\subsection{Setup}
\label{sec:evaluation:setup}

To conduct our evaluation experiments, we implemented the proposed approach as 
a plug-in for the Java-based open-source Process Mining Framework 
\textit{ProM~6}.\footnote{See \url{http://www.promtools.org/}}

\smallskip
\noindent \textbf{Behavioral Models.} Using the implementation and the datasets 
described above, we conducted experiments with the following behavioral models:
\begin{compactitem}
	\item \textbf{TE:} The \textit{trace equivalence model} from \autoref{sec:traceequivalence}.
	
	\item \textbf{2G, 3G, 4G:} The \textit{N-gram model} from 
	\autoref{sec:ngram}, using $N=2$ (the strictest notion), $N=3$, and  $N=4$, respectively.  
	\item \textbf{WO:} The \textit{weak order model} from \autoref{sec:weakorder}. 
\end{compactitem}

\noindent
Note that when a behavioral model returns a probability of zero for all 
possible resolutions of a trace, which happens if the model cannot derive any 
evidence from the event log, we regard all resolutions to be equally likely. 
That is, we assign a uniform probability $|\Phi(\sigma)|^{-1}$ to each 
resolution.    

\smallskip 
\noindent \textbf{Approximation methods.} To evaluate the impact of our proposed method for result approximation, we compute results based on two configurations:
\begin{compactitem}
\item \textit{No approximation}: the conformance for all resolutions with a non-zero probability is assessed.

\item \textit{$\alpha$=0.99 approximation}: the approximation method from \autoref{sec:resultapproximation} with $\alpha=0.99$ and $\delta=0.10$. 
%
\end{compactitem}

\smallskip 
\noindent \textbf{Performance measures.} To determine how accurate the obtained conformance values are, we compare them to the true conformance values, i.e., a gold standard value, based on the order in which a trace's events appear in a log. For the purposes of this evaluation, we consider conformance in terms of (weighted) fitness, as defined in \autoref{sec:partialorderresolution}. 
We consider accuracy at both the trace- and the log-levels, where the former 
focuses on the accuracy for individual traces and the latter on the accuracy 
obtained per log.

At the trace-level, we quantify the difference between \textit{obtained}  and 
\textit{true} fitness values by employing the Root-Mean Squared Error (RMSE). 
Given an event log $L$, a process model $M$, and a behavioral model $B$, the 
RMSE is given as follows:

\begin{equation}
\label{eq:rmse}
\mathit{error_{tracelevel}}(L, M, B)=  \sqrt{\frac{\sum_{\sigma \in 
L \setminus \Sigma_{certain}}(\mathit{fit}(\sigma, M)  - P_{\mathit{fit}}^B(\sigma, M))^2}{| L \setminus \Sigma_{certain}|} }
\end{equation}

At the log-level, we quantify the difference between the \emph{obtained} and 
\emph{true} fitness values by computing the absolute difference:

\begin{equation}
\label{eq:logerror}
\mathit{error_{loglevel}}(L, M, B)=  | \mathit{fit}(L, M) -  
P_{\mathit{fit}}^B(L, M) |
\end{equation}

Note that, as shown, the trace-level results are computed over only the traces 
with uncertainty, i.e., $L \setminus \Sigma_{certain}$, whereas the log-level 
results are computed over all traces in $L$.

\smallskip
\noindent \textbf{Baselines.} As a basis for comparison, we employ two baseline techniques, BL1 and BL2: 
\begin{compactitem}
	\item \textbf{BL1:} This baseline follows state-of-the-art work by  considering each potential resolution of an uncertain trace to be 
	equally likely, such as proposed by Lu et al.~\cite{lu2014conformance}. As such, the baseline 	assigns  	a \textit{uniform probability} of $|\Phi(\sigma)|^{-1}$ to each resolution. The 	comparison against this baseline is intended to demonstrate the value of using 	probabilistic models that assess the likelihood of different resolutions.
	
	\item \textbf{BL2:} This baseline deals with order uncertainty by simply 
	excluding all traces that are affected by it, i.e., basing the computation of 
	log fitness on only those traces without any order uncertainty. Naturally, 
	this baseline can only be used to assess accuracy at the log-level because it 
	cannot compute results for traces with order uncertainty.
\end{compactitem}

\subsection{Results}
\label{sec:evaluation:results}

This section presents results on the accuracy of conformance checking using the 
proposed behavioral models for various noise levels and degrees of uncertainty (\autoref{sec:accuracy}), followed by analysis of the computational 
efficiency and accuracy of the approximation method (\autoref{sec:efficiency}).

\subsubsection{Result Accuracy}
	\label{sec:accuracy}

\begin{figure}[!htb]
	\centering
	\begin{minipage}{0.47\textwidth}
		\includegraphics[width=1.0\textwidth]{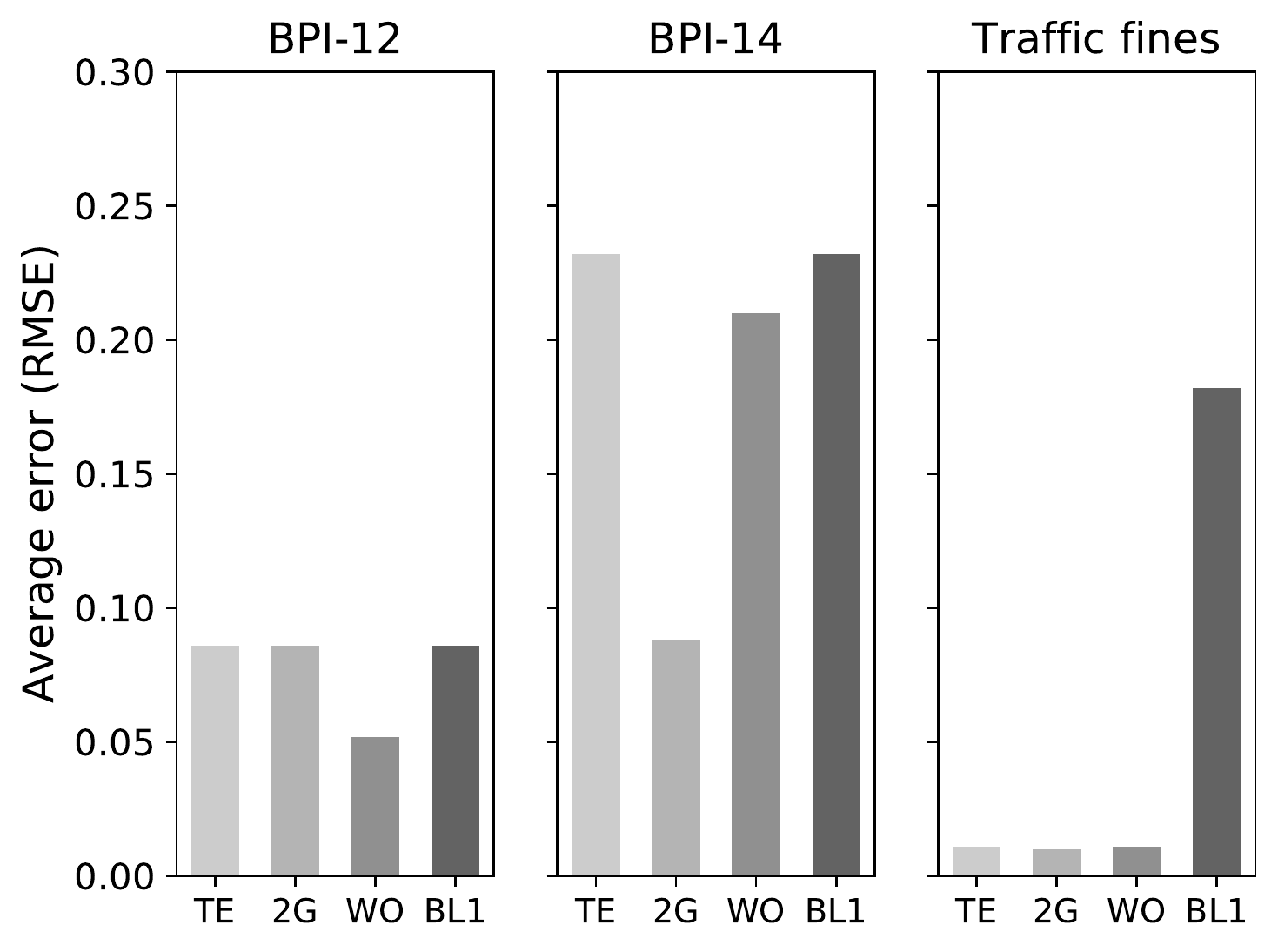}
		\caption{Results accuracy real-world logs (trace-level)}
		\label{fig:realworld_acc_tracelevel}
	\end{minipage}
	\hfill
	\begin{minipage}{0.47\textwidth}
			\includegraphics[width=1.0\textwidth]{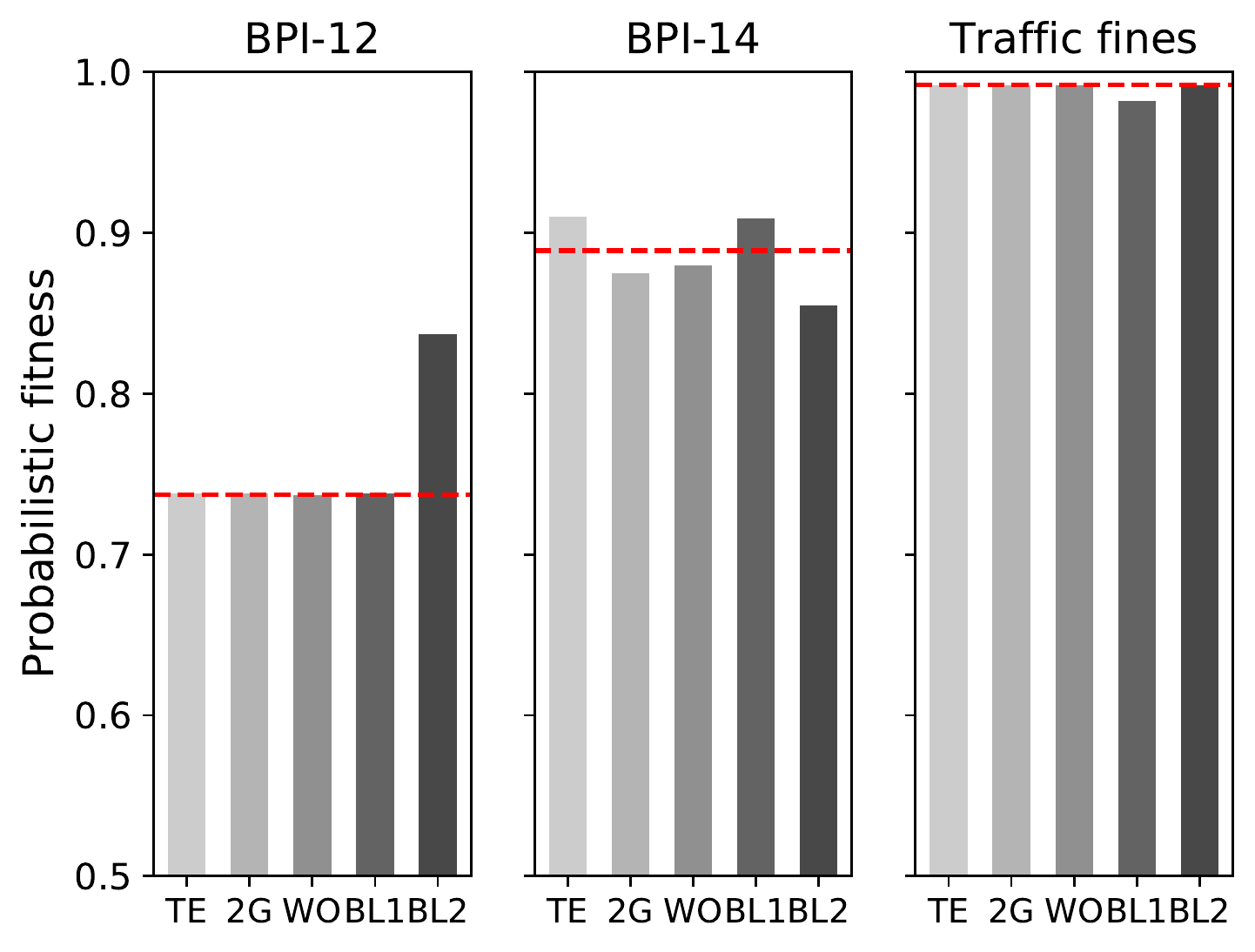}
		\caption{Results accuracy real-world logs (log-level)}
		\label{fig:realworld_acc_loglevel}
	\end{minipage}
\end{figure}

\mypar{Real-world collection} Figures~\ref{fig:realworld_acc_tracelevel} and~\ref{fig:realworld_acc_loglevel}
visualize the conformance checking accuracy obtained by the behavioral models and baselines 
for the real-world event logs.\footnote{Note that, for clarity, we only depict 
the best performing N-gram model, i.e., $2G$.}  In general, the figures reveal 
that the result accuracy on the trace-level is subject to much more variance 
than the result accuracy on the log-level. While the trace-level RMSE 
(\autoref{fig:realworld_acc_tracelevel}) differs quite considerably across the 
different behavioral models, the log-level error is relatively stable 
(\autoref{fig:realworld_acc_loglevel}). Note that the red dashed line in 
\autoref{fig:realworld_acc_loglevel} shows the true fitness value.  

Taking a look at the details of the trace-level results, we see that the weak 
order model (WO) outperforms the others for the BPI-12 log (RMSE of 0.052 
versus 0.086). However, for the BPI-14 log, the 2G model performs much better 
than the WO model (RMSE of 0.088 versus 0.210). For the Traffic fines log, the 
behavioral models achieve the same overall performance. Noticeably, the best 
performing models consistently outperform the baseline approach BL1. The RMSE 
of the Traffic fines case primarily shows that the uniform probability 
distribution  from BL1 can result in a a considerably reduced trace-level 
accuracy (RMSE of 0.182 versus 0.011 of the proposed models). 

The details of the log-level results show that all behavioral models have the 
same accuracy for the BPI-12 and the Traffic fines log. For the BPI-14 log, we 
observe slightly more variation. Here, the WO model is closest to the actual 
fitness value. In general, we see that the behavioral models consistently 
perform equally or better than both baselines. However, BL1 is more accurate 
than BL2, which is considerably off for the BPI-12 and BPI-14 
logs, given their high amounts of uncertain traces (38.3\% and 93.4\%, 
respectively). Nevertheless, it is interesting to recognize that the fitness 
for the BPI-14 log is better than perhaps expected for an approach that is only 
able to consider less than 7\% of the total traces in an event log. 

\begin{figure}[!htb]
	\centering
	\begin{minipage}{0.47\textwidth}
		\includegraphics[width=1.0\textwidth]{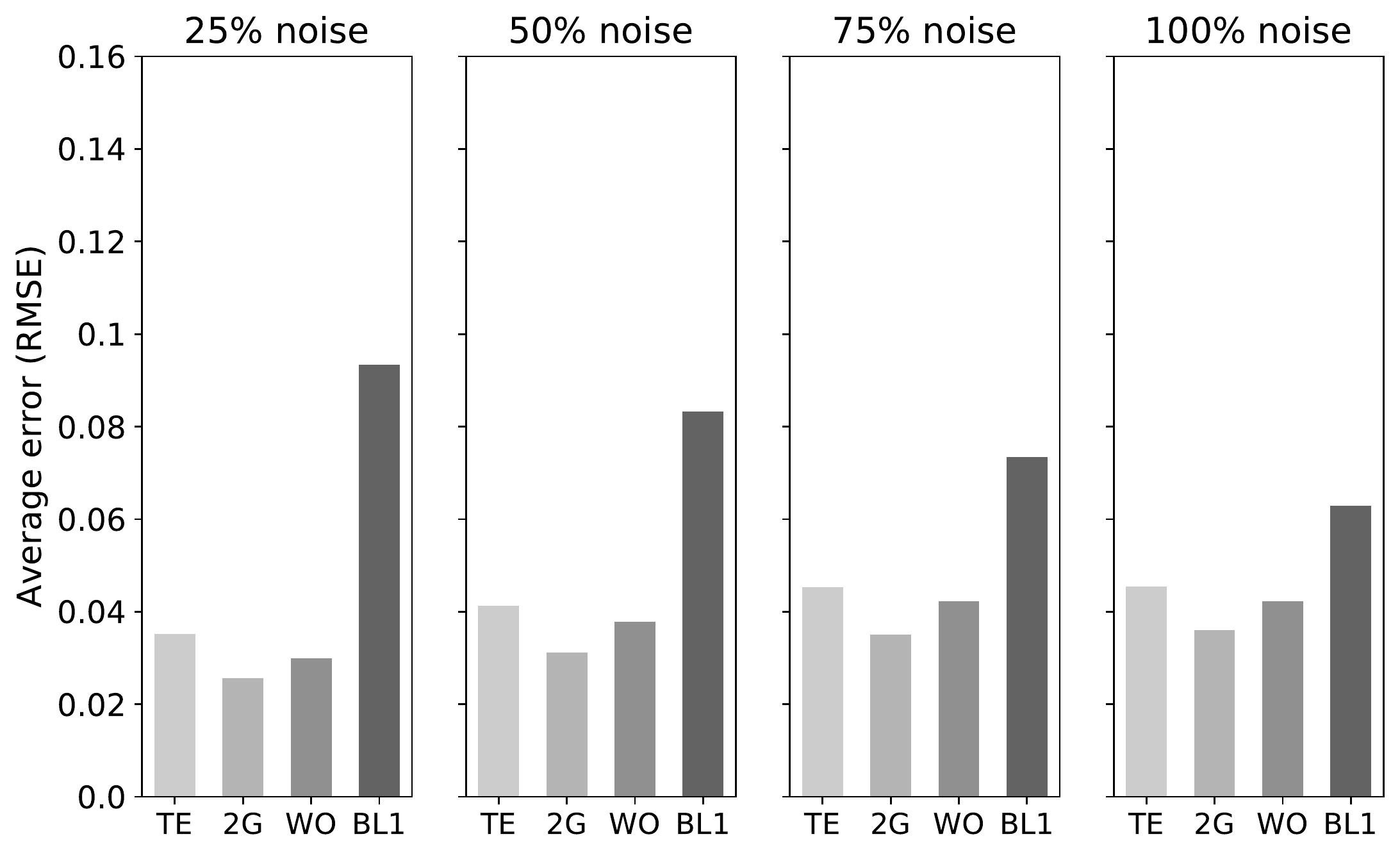}
		\caption{Results accuracy synthetic logs (trace-level)}
		\label{fig:generated_acc_tracelevel}
	\end{minipage}
	\hfill
	\begin{minipage}{0.47\textwidth}
		\includegraphics[width=1.0\textwidth]{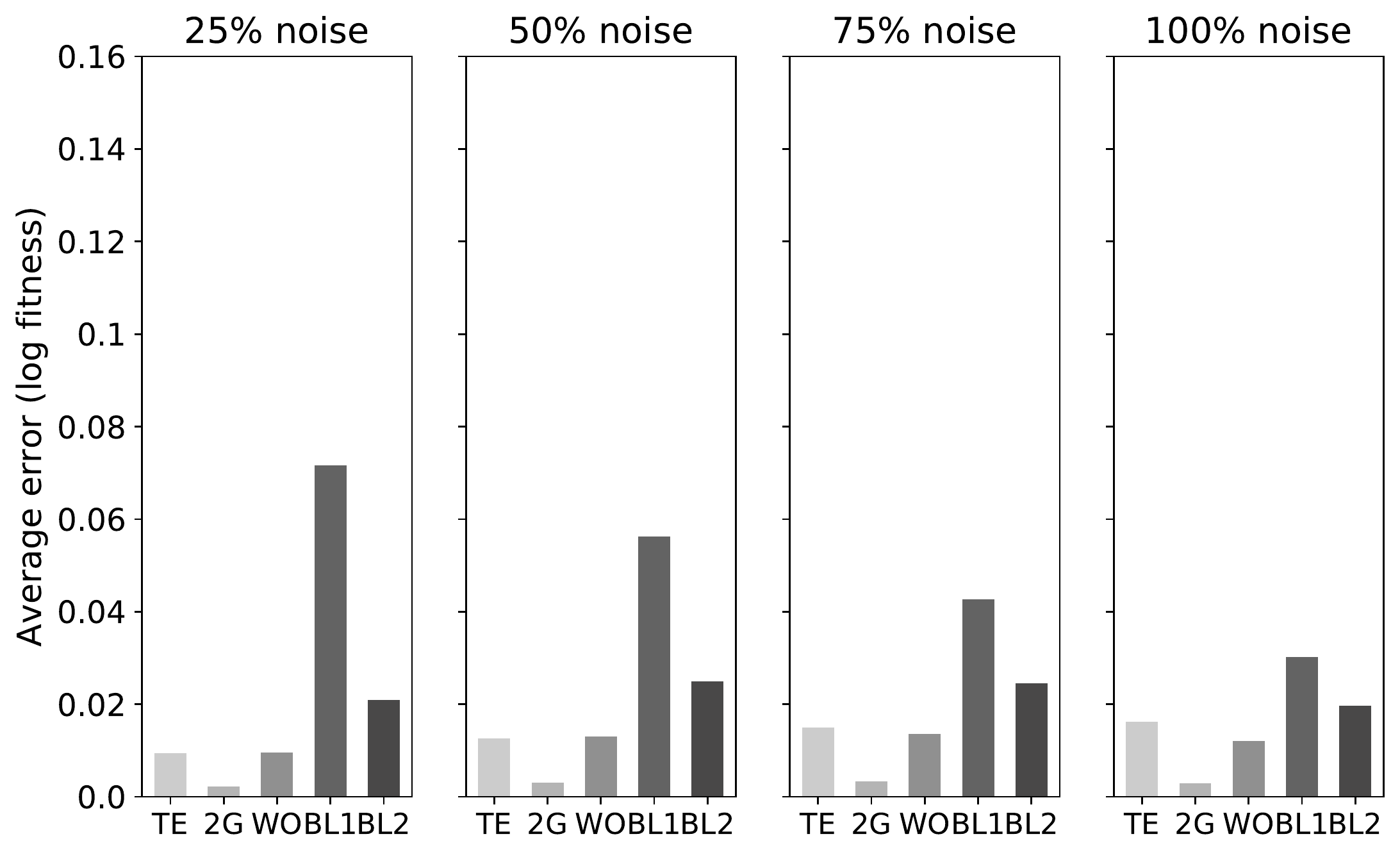}
		\caption{Results accuracy synthetic logs (log-level)}
		\label{fig:generated_acc_loglevel}
	\end{minipage}
\end{figure}

\mypar{Synthetic collection} 
Figures~\ref{fig:generated_acc_tracelevel}~and~\ref{fig:generated_acc_loglevel} visualize the results obtained for the 500 generated process models, over varying noise levels.
The figures clearly show that the general trends of the results are preserved 
across noise levels and apply to both trace and log-level error measurements. 
In particular, the 2-gram model (2G) here consistently outperforms the other models, achieving a trace-level RMSE ranging from 0.026 (25\% noise) to 0.036 (100\% noise) and a log-level error between 0.002 and 0.003. 
At the trace-level, the 2G model is closely followed by the WO model, with an 
RMSE between 0.030 and 0.042, though \autoref{fig:generated_acc_loglevel} 
clearly reveals a larger difference when considering the log-level. There, the 
WO model still performs second best, but achieves an error between 0.010 and 
0.012, i.e., between 4 and 5 times as large as the error of the 2G model. 

 When aggregating the accuracy of the behavioral 
models over all noise levels, we observe that the 2G model 
achieves an average RMSE of 0.032 and a log-level fitness error of 0.003.
By contrast, the WO model model achieves an RMSE of 0.038 (1.2 times as high) and a log-level error of 0.013 (4.3 times as high as the 2G model).
The trace equivalence model (TE) performs worst, with an RMSE of 0.042 (1.3 times) and log-level error of 0.015 (4.9 times).
Nevertheless, as also depicted, all models considerably outperform both 
baselines. The uniform probability baseline (BL1) has the worst performance, 
obtaining an RMSE of 0.078 (2.40 times the error of 2G) and a log-level error 
of 0.043 (14.3 times as high). BL2, which simply ignored all traces with 
ordering uncertainty, achieves a log-level error of 0.023 (7.8 times the error 
of 2G). 

Considering the tracel-level results obtained per event log, we observe some variability among the performance of the behavioral models. 
Out of the 2000 cases (500 models over 4 noise levels), the 2G model outperforms all other models in the majority of the cases: 1203 times. Surprisingly, the trace equivalence model (TE) performs best for 533 cases, while the weak order model (WO) outperforms the others in 264 cases.
There is no event log in the synthetic collection for which the baselines 
outperform the proposed models. 

\mypar{Impact of order uncertainty}
Aside from assessing the accuracy for different noise levels, it is interesting to assess how the degree of uncertainty in an event log affects the conformance checking accuracy.
To obtain event logs that cover a broad range of uncertainty degrees, we generated additional event logs for the models in the synthetic data collection using different throughput rates. In particular, for each of the 500 models, we also generated event logs with 0.25, 0.50, and  2.0 times the throughput rate of the event logs used in the experiments described earlier. The idea here is that a higher (lower) throughput rate results in more (fewer) events that have a timestamp within the same minute and, therefore, more (fewer) uncertainty. 

We grouped the obtained event logs based on their percentage of traces with order uncertainty, as shown in \autoref{tab:acc_uncertaintylevels}.
The table reveals the overall expected trend in which the average  error increases along with the amount of order uncertainty present in an event log. However, it is interesting to observe that this applies in only a very limited manner to the 2G model, which is both the best performing and most stable model across the various degrees of uncertainty. By contrast, the performance of the TE model quickly decreases with higher uncertainty. This makes sense, given that this model requires traces without any form of uncertainty in order to be able to derive evidence for its probability computation.
Although it is more stable than TE, the WO model is outperformed by the 2G model. Finally, the results clearly show that each of the proposed models outperform the baselines. In line with expectations, especially the performance of BL2, which only consider traces without uncertainty, sharply decreases for higher amounts of uncertainty.

\begin{table}[!htb]
	\centering
	\small
	\begin{tabular}{lcccccccc}
		\toprule
		\textbf{Uncertain trace ratio} & \textbf{0.0--0.2} & \textbf{0.2--0.4}& \textbf{0.4--0.6}& \textbf{0.6--0.8}& \textbf{0.8--1.0}\\
		Number of event logs & 304 &1110 & 877 & 539 & 170 \\
		Uncertain events (avg.) & 7.0\% & 13.2\% & 19.5\% & 25.5\% & 36.2\% \\
		\midrule 
		TE & \textbf{.001}& \textbf{.004} &.011 &.020 & .033 \\
		2G & \textbf{.001} &\textbf{.004} & \textbf{.002} & \textbf{.002} &\textbf{.004} \\
		\smallskip
		WO & .002 & .010 & .012 & .011 & .016 \\
		BL1 & .006 & .038 & .046 & .044 & .049 \\
		BL2 &  .018 & .020& .027 & .033 & .082 \\
		\bottomrule
	\end{tabular}
	\caption{Log-level error for different levels of uncertainty in synthetic logs}
	\label{tab:acc_uncertaintylevels}
\end{table}

\subsubsection{Computational Efficiency}
		\label{sec:efficiency}
		
		To assess the runtime efficiency of conformance checking under uncertainty and our proposed approximation method, we conducted experiments on a 2017 MacBook Pro (Dual-Core Intel i5) with 3,3GHz and an 8GB Java Virtual Machine.

\smallskip
\noindent\textbf{Real-world collection.} 
The results obtained by applying the approach without and with approximation on the three real-world event logs are shown in \autoref{tab:runtimerealworld}. We here report on the runtimes obtained using the 2G behavioral model, which achieves the overall best performance accuracy.
The recorded runtimes highlight that conformance checking in the presence of traces with order uncertainty can take considerable time. Although the Traffic fines log is processed in less than half a second, the BPI-14 log requires almost 40 minutes, whereas the BPI-12 log takes close to 4 hours to process. 
These lengthy runtimes follow from the combination of various factors:
the traffic fines log is has short traces (average length of 3.7 events), a low 
number of variants (231), and a relatively low amount of uncertainty. This 
means that the vast majority of its 150,370 traces do not require intensive 
expensive alignment computation, given that they do not show uncertainty and 
follow a previously observed trace variant.
By contrast, the BPI-12 and BPI-14 event logs have longer traces (20.0 and 7.3 
events per trace, respectively), have more variants (4,366 and 31,725) and a 
higher fraction of uncertainty (38.3\% and 93.4\% of the traces). The runtime 
difference between the BPI-12 and BPI-14 logs can, most likely, be attributed 
to the average length of traces, which is considerably higher for the BPI-12 
case (20.0 versus 7.3 average events per trace).

When considering the results including our proposed approximation method, 
first and foremost, we note that approximation has minimal impact on the 
conformance checking accuracy. For all real-world logs, the (trace-level) error values obtained when 
using our approximation method are equal up to 3 decimal places compared to those obtained without using approximation. Our results cover a configuration 
with $\alpha = 0.99$. Yet, experiments with $\alpha = 0.95$ yielded 
near-identical results.

This near-identical accuracy is obtained while potentially achieving considerable improvements in terms of runtime.
As shown in \autoref{tab:runtimerealworld}, approximation does not impact the Traffic fines event log. Due to the relatively low order uncertainty in the event log and short trace length, there are no traces for which there are sufficient possible resolutions to trigger the approximation approach. However, we do observe considerable gains for the other two event logs. In particular for the BPI-12 event log, we observed that the approximation method nearly halves the time required to obtain conformance checking results (gaining 47.9\% of the runtime). For the BPI-14 log, the runtime improvement is smaller though still notable, with 13.5\% of the runtime saved. These results imply that the benefits of the approximation methods are particularly apparent when they are needed most, i.e., for conformance checking settings with otherwise high runtimes. This conclusion is strengthened by also considering the results obtained for the synthetic data collection.

\begin{table}
	\centering
	\small
	\caption{Runtime efficiency on the real-world  logs.}
	\label{tab:runtimerealworld}
	\begin{tabular}{lccc}
		\toprule
		\textbf{Measure} & 	   \textbf{BPI-12} & \textbf{BPI-14} & \textbf{Traffic fines} \\
		\midrule 
		Runtime (no approx. )&    237.8m & 39.6m & 424ms \\
		Runtime (with approx.) & 123.8m & 34.2m & 390ms\\

	Time saved through approx. & 114.0m (47.9\%) & 5.4m (13.5\%) & n/a \\

		Traces approximated & 1,090 (8.3\%)& 2,318 (5.6\%)& 0 (0.0\%) \\

		Additional error (RMSE) & 0.000 & 0.000 & 0.000 \\
		\bottomrule
	\end{tabular}	
\end{table}

\smallskip
\noindent\textbf{Synthetic collection.} \autoref{tab:runtimesynthetic} presents 
the runtime results obtained for the synthetic data collection. We observe that the approximation method achieves 
considerable time savings, averaging between 58.9\% and 45.9\% savings per log 
and a maximum of 85.7\% time saved. These gains  are achieved while having a 
limited impact on the conformance checking accuracy of the approach. On 
average, the additional error incurred based on approximation is between only 
0.02\% and 0.03\%, whereas the maximum additional error is 1.0\%.

It is interesting to observe that the approximation method is actually applied to only about 2.0\% of the traces with uncertainty, with a maximum of 18.8\% in a single log. However, as shown by the much larger gains in runtime, it is clear that the approximation method is applied to traces that contribute the most to the total runtime. This naturally follows from the approximation method's dependence on the central limit theorem, which enforces that approximation can only be applied to traces with at least 20 possible resolutions. In other words, the approximation method is generally applied to traces with the otherwise largest runtime.

	\begin{table}[!hbt]
	\centering
	\small
	\setlength{\tabcolsep}{0.5em}
	\caption{Runtime efficiency on the synthetic collection (averages over 
		collection).}
	\label{tab:runtimesynthetic}
	\begin{tabular}{lrrrrr}
		\toprule
		& \multicolumn{4}{c}{\textbf{Noise level}} \\
		\textbf{Measure} & \textbf{25.0} & \textbf{50.0} & \textbf{75.0} & \textbf{100.0} \\
		\midrule 
		Time, no approx. (avg.) & 12.1s & 12.8 & 13.8s & 14.0s \\
		\smallskip
		Time, with approx. (avg.) & 5.0s & 6.0s & 6.7s & 7.6s \\
		
		Time saved through approx. (avg.)& 58.9\% & 53.1\% & 51.2\% & 45.9\% \\
		\smallskip
		Time saved through approx. (max.)& 85.7\% & 80.9\% & 84.6\% & 80.7\% \\
		
		Traces approximated (avg.) & 1.7\% & 1.9\% & 2.0\% & 2.2\% \\
		\smallskip
		Traces approximated (max.) & 17.2\% & 18.7\% & 17.9\% & 18.8\%\\
		Approx. error (avg.) & 0.03\% & 0.02\% & 0.03\% & 0.02\% \\
		Approx. error (max.) & 1.00\% & 0.98\% & 1.01\% & 1.00\% \\
		\bottomrule		
	\end{tabular}
\end{table}

\subsection{Discussion}
	\label{sec:discussion}

	The results presented in this section show that the 2G model is generally the best performing one, though there are cases when the WO model and, occasionally  the TE model achieve the best accuracy.
	A factor that plays an important role for the performance of the behavioral models is the amount of information that the model can utilize from the available event log. The higher the abstraction level of a behavioral model, the more information that can be taken into account. The trace equivalence model, the least abstract model, can only derive probabilistic information from traces that do not contain any uncertainty. Even if such traces are available, the model can only provide probabilistic insights if the same trace is repeated for the process. This makes the trace equivalence model inapplicable to logs with a high variety and uncertainty, such as the BPI-14 log, in which over 93\% of the traces have uncertainty. This problem can also occur for the 4G and 3G models, which still depend on repeated sequences of events without uncertainty.  
	
	\mypar{Best performing behavioral model}
	Overall, we conclude that the 2G model performs best.
	Especially on the widely varying synthetic model collection, the 2G model is shown to achieve the most accurate trace-level and log-level results across all noise levels (Figures~\ref{fig:generated_acc_tracelevel} and~\ref{fig:generated_acc_loglevel}) and degrees of uncertainty (\autoref{tab:acc_uncertaintylevels}). However, the WO model outperforms the others for the BPI-12 log. This may be attributed to the larger average length of the traces, which may suggest, that in this case, it is more helpful to consider events that indirectly follow each other, rather than only those that directly follow each other.
	Note that a post-hoc evaluation of the results did not reveal any notable correlation between process model characteristics, such as the presence of loops or non-free choice components, and the accuracy obtained by the different behavioral models.
	
	\mypar{Impact of approximation}
	Aside from the selection of a behavioral model, the results from \autoref{sec:efficiency} show that the proposed approximation method can be used with little impact on the approach's accuracy, while gaining considerable benefits in terms of runtime. The preservation of such a high level of result accuracy is due to the method's grounding in statistical distributions, which requires that at least 20 possible resolutions of a  trace are checked before applying approximation.
	
	\mypar{Applicability of the approach}
	As defined in our event model presented in \autoref{sec:preliminaries}, our approach  targets cases in which order uncertainty is explicitly captured through the presence of events with identical timestamps. This is a situation that has clearly been shown to be prevalent in practical settings through our analysis of the real-world event logs, as depicted in \autoref{tab:realworldcollectionCharacteristics}. 
	However, we recognize that there are also cases in which order uncertainty exists, but may not be directly visible from the granularity of the available timestamps. For instance, an unreliable sensor may record timestamps in terms of milliseconds, even though the true moment of occurrence can only be guaranteed in terms of seconds. These cases could be detected using approaches such as proposed by Dixit et al.~\cite{dixit2018detection}. Following such a detection, a pre-processing step could be employed that abstracts the unreliable timestamps to a granularity at which the uncertainty no longer exists, before employing our proposed approach. In this manner, also such occurrences of order uncertainty can be detected, while maintaining the benefits of our approach in comparison to approaches that impose strict assumptions on the correct order of a trace.

\section{Related Work}
\label{sec:relatedwork}
In this section we discuss the three primary streams to which our work relates: conformance checking, sequence classification, and uncertain data management.

\textit{Conformance checking} is commonly based on alignments, as discussed 
 in \autoref{sec:conformancecheckingBG}. 
Due to the computational complexity of alignment-based conformance checking, various angles have been 
followed to improve its runtime performance. Efficiency improvements 
have been obtained through the use of search-based 
methods~\cite{DBLP:conf/otm/ReissnerCDRA17,DBLP:conf/bpm/Dongen18}, and planning 
algorithms~\cite{DBLP:conf/sebd/LeoniM17}. 
Similar to the approximation method presented in \autoref{sec:resultapproximation}, several approaches approximate conformance results to gain 
efficiency, e.g., 
by employing approximate alignments~\cite{DBLP:conf/bpm/TaymouriC16}, sampling strategies~\cite{bauer2019estimating}, and 
applying divide-and-conquer schemes in the computation of conformance 
results~\cite{AalstV14,munoz2014single,
	DBLP:journals/sosym/LeemansFA18}.
These approaches can be regarded as complimentary to our approximation method, since our method reduces the number of resolutions for which conformance results need to be obtained, whereas these other approaches improve the efficiency achieved per resolution.

\textit{Sequence classification} refers to the task of assigning class labels 
to sequences of events~\cite{xing2010brief}, a task with applications, e.g., in 
genomic analysis, information retrieval, and anomaly detection. A key challenge 
of sequence classification is 
the high dimensionality of features, resulting from the sequential nature of 
events. A broad variety of methods have been developed for this task, 
which can be generally grouped over three categories~\cite{xing2010brief}, i.e., \emph{feature-based}, \emph{distance-based}, and \emph{model-based} methods. In particular the latter methods, which employ techniques such as Hidden Markov Models, may resemble the behavioral 
models proposed in this work. However, the addressed problem fundamentally differs: Sequence classification 
assigns labels to sequences, whereas our models aim at determining 
the actual sequence of events itself.


\emph{Data uncertainty} is inherent to various application contexts, typically 
caused by  data randomness, incompleteness, or limitations of measuring 
equipment~\cite{pei2007probabilistic}. This has created a need for algorithms 
and applications for uncertain data management~\cite{aggarwal2009survey}.  These include models for probabilistic and 
uncertain databases, see~\cite{peng2015supporting}, along with 
respective query 
mechanisms~\cite{murthy2011making}. 
Moreover, models to capture uncertain event occurrence. This includes models 
where the event occurrence itself is 
uncertain~\cite{DBLP:conf/sigmod/ReLBS08} 
as well as those where 
uncertainty relates only to the time of event 
occurrence~\cite{DBLP:journals/is/ZhangDI13}, which may be defined by a density 
function over an interval. Our model, in turn, incorporates the particular 
aspect of order uncertainty within a trace. The reason being 
that the probability of an event occurrence at a particular time is of 
minor importance when assessing conformance of a trace with respect to the 
execution sequences of a model. Aside from the order uncertainty we consider, uncertainty can also follow from unknown mappings of events to process model activities~\cite{vanderaa2019efficient,baier2014bridging} and from the use of ambiguous process specifications~\cite{vanderaa2018checking}.

\section{Conclusion}
\label{sec:conclusion}

In this paper, we overcame a key assumption of conformance checking that 
limits its applicability in practical situations: The requirement that the 
events observed for a case in a process are totally ordered. In particular, we 
established various behavioral models, each incorporating a different notion of behavioral abstraction. These behavioral models enable the resolution of partial 
orders by inferring probabilistic information from other process executions. 
Our evaluation of the approach based on real-world and synthetic data reveal that our approach  achieves considerably more accurate results than an existing baseline, reducing the average error by 59\%. 
To cope with the runtime complexity of conformance checking with order uncertainty, we also presented a sample-based approximation method. 
The conducted experiments demonstrate that this method can lead to considerable runtime reductions, while still obtaining a near-identical conformance checking accuracy.

In our experiments, we focused on conformance on the trace and log level. However, as 
indicated in \autoref{sec:partialorderresolution}, the assignment of 
probabilities to resolutions is also beneficial for advanced 
feedback on non-conformance. Measures that quantify conformance may be computed 
per resolution and then be aggregated based on the assigned probabilities. 
Similarly, our probabilistic model highlights the overall importance of particular 
deviations. 
However, we see open research questions related to the perception of such 
probabilistic results in conformance checking and intend to explore this aspect 
in future case studies. We also aim to extend our work with 
behavioral models that go beyond temporal aspects of event logs. For instance, 
employing decision mining techniques, multi-perspective models may enable 
more fine-granular selection of traces for partial order resolution.

\noindent \textit{Reproducibility: Links to the employed source code and datasets  are given in \autoref{sec:evaluation}.}

\section*{Acknowledgment}
\vspace{-1em}
Part of this work was funded by the Alexander von Humboldt Foundation.

\bibliographystyle{elsarticle-num}
\bibliography{ref}

\end{document}